# Decision Trees for Helpdesk Advisor Graphs

S. Gezerlis and D. Kalles

*Abstract*—We use decision trees to build a helpdesk agent reference network to facilitate the on-the-job advising of junior or less experienced staff on how to better address telecommunication customer fault reports. Such reports generate field measurements and remote measurements which, when coupled with location data and client attributes, and fused with organization-level statistics, can produce models of how support should be provided. Beyond decision support, these models can help identify staff who can act as advisors, based on the quality, consistency and predictability of dealing with complex troubleshooting reports. Advisor staff models are then used to guide less experienced staff in their decision making; thus, we advocate the deployment of a simple mechanism which exploits the availability of staff with a sound track record at the helpdesk to act as dormant tutors.

## I. Introduction

Customer satisfaction is a key factor in making clients loyal to a service provider [1]. In telecommunications, this is closely linked to the time it takes to fix a problem and the way a fault request is handled. Long-term customers are considered more loyal to a brand and usually cost less to serve [2][3].

Big telecommunication organizations have adopted Standard Operating Procedures (SOPs) and Key Performance Indicators (KPIs) to help guide their policies in servicing customer fault complaints.

SOPs are guidelines to be followed by engineers, to help them decide how to solve a particular problem. SOPs are usually combinations of rules and decision trees [4][5] and constitute an integral part of staff training.

KPIs are numeric indices which attempt to capture aspects of service quality (as perceived by the customer or by the organization) and are usually expressed in terms of average request handling speed, percentage of problems solved remotely, or repeated complaint rates. Though a KPI may be straightforward to define and to measure, relating it to soft attributes of service provision is an elusive task. For example, being able to conclude a job on time is an obvious target for a maintenance engineer, and so is the ability to draw as few organizational resources as possible for any particular task. Though relating both to cost can also be straightforward, linking them to measures of customer satisfaction can be a difficult exercise. For this reason, composite quality indices are notoriously difficult to define and are subject to painstaking secrecy and to simplifications [6].

SOPs and KPIs are usually deployed at two levels of support: Remote (phone) Support, which handles all incoming complaints and calls, and On-Site Service with field technicians, who work on the landline network. The decision to deploy On-site Support while already providing Remote Support is sometimes a function of the Remote Support technician's experience, workload, mental situation and agility and either approach has to be judged based on the average cost each of them incurs on the company, all of which are also functions of time.

The element of cost is captured by the Operational Expense (OPEX) formulation, where one assumes that a complaint remotely resolved costs less, is resolved faster and still keeps customer experience at a satisfactory level. With $OPEX_{OS}$ standing for OPEX for On Site Support at 1, $OPEX_{RS}$ for Remote resolution Service is $OPEX_{OS}/n$, with typical values of $n > 10$ (details are business confidential).

The complexity of making support-related decisions on time, with pressure and with acceptable consistency with organizational policies on customer support and cost containment, cannot be overstated. As training is a key to improvement, this paper puts forward an unconventional approach, based on the on-the-fly identification of key personnel who can temporary act as advisors to their peers in a discreet, non-intrusive fashion.

The contribution of this article is two-fold. First, we use decision trees to identify inconsistencies in the cost incurred when dealing with customer complaints; we interpret such inconsistencies as a measure of the ability of the company to deliver an as-uniform-as-possible customer experience in complaint troubleshooting. Then, we use these inconsistencies to rank helpdesk agents who troubleshoot customer complaints; this generates a knowledge hierarchy that is dynamically updated based on identifying personnel who can advise or who need to be advised.

The work serves as a bridge between the formality of logic that is inherent in expressing SOPs and the inherent vagueness as regards what is the "best" way to resolve a particular customer complaint. It also advances the state of the practice in the customer relationship management (CRM) genre [7], where what little has been done until recently [8][9][10] as regards customer complaints, mainly focuses on knowledge about the value of the client (instead of complaint resolution).

## II. DECISION TREES AND ADVISOR FLOW GRAPHS

For our case study, we selected a small group of Remote Support agents (technicians), who operated under the supervision of one of the co-authors, and recorded their measurements as attributes for Internet and TV complaints instances. Each instance corresponds to an Internet or IPTV (Internet Protocol TV) over-xDSL complaint, where we record synchronization between the Integrated Access Device and the Central Office routers, in addition to other copper-related metrics. For these agents, we also recorded whether they handled the complaint at the helpdesk level or referred it to On Site Support. One aims at maximizing the number of issues resolved remotely, without having to resort to On Site Support. Though this makes obvious sense from a financial point of view, it is not trivial; sub-standard problem resolution generates recurring complaints, which may or may not be dealt by the same agent next time, thus raising costs (which is seldom the case for the more expensive On Site Support).

The training set we use to build a decision tree, using the C4.5 algorithm [4], looks like the one shown in Table I.

TABLE I
A SNAPSHOT OF A TRAINING SET

| Agent | Product | Area | Profile | Sync | Max | Dist | State |
|---|---|---|---|---|---|---|---|
| … | … | … | … | … | … | … | … |
| AGENT04 | INTERNET | Athens | 2 | 2048 | 9138 | 2.7 | OS |
| AGENT02 | INTERNET | Patras | 30 | 26421 | 27786 | 1.8 | RS |
| AGENT03 | INTERNET | Athens | 24 | 2045 | 3408 | 3.1 | OS |
| ... | … | … | … | … | … | … | … |

*State* is the class variable (showing the type of support, On-Site/Remote).
*Product* refers to the service a client has bought and for which service the complaint is being lodged.
*Area* refers to the city where the fault occured.
*Profile* refers to the nominal speed of the connection for that client (not the actual capacity of the line, but the service capacity; for example, a low-cost client might buy a 2 Mb/s service even though the line might accommodate much larger speeds).
*Sync* refers to the actual synchronization speed of a client's line.
*Max* refers to the theoretical maximum attainable synchronization of a client's line based on signal strength, attenuation, and a variety of physical characteristics.
*Dist* refers to the copper cable length from the central circuit infrastructure (measured in km).

The training set was made up of about 1,500 helpdesk reports, recorded over a period of about 3 weeks, by automatically logging data collected at distributed corporate information systems, for 5 helpdesk agents. The integrated data set has been generated by combining those data together.

### A. Cost-oblivious decision trees

For the base-level experiment we treat individual agents as attributes. We have used RWeka and R for analyzing the data and building the models. Running a 10-fold cross-validation delivered an accuracy of 69.67% of predicting whether the complaint would be resolved at the on-site or at the remote level. Fig. 1 shows a snapshot of the decision tree; the (dark) multi-valued AGENT attribute appears at all nodes near the tree root and, clearly, should be avoided in any (agent-agnostic) SOP. Dropping that attribute along with the AREA attribute (agents service requests from a central location which may better resolved just by knowing singularity aspects of the actual physical network) resulted in an accuracy of 60.65%.

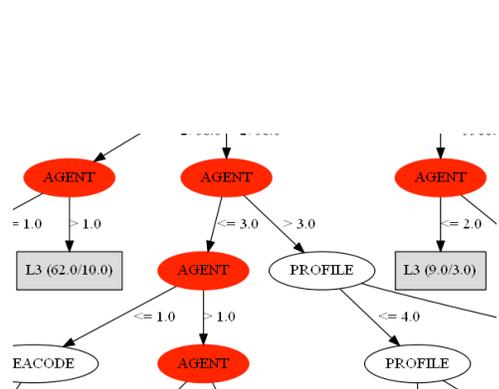

Fig. 1. A decision tree using all data labels - *agent* as a decision node.

We then use the data sets of individual agents to develop decision trees which capture how each agent treats a customer complaint as well as a measure of treatment uniformity. The measurement is based on partitioned data sets which are cross-tested [11]. We partition the overall complaints data set into groups, according to how complaints were assigned to agents and examine whether the problem resolution practices of one agent apply to customer complaints handled by another agent. On one hand, this approach allows us to see how each agent's decision model is (or is not) compatible with other agents' models. On the other hand, however, it also allows us to consider model differences compared not to a theoretical maximum of 1 but to a practical maximum as calculated by the re-classification error of each agent's model.

TABLE II
CORRECTLY CLASSIFIED INSTANCES - CROSS-TEST RESULTS

|   | 1 | 2 | 3 | 4 | 5 |
|---|---|---|---|---|---|
| 1 | **0.60** | 0.38 | 0.48 | 0.54 | 0.50 |
| 2 | 0.49 | **0.72** | 0.56 | 0.50 | 0.42 |
| 3 | 0.49 | 0.66 | **0.60** | 0.50 | 0.42 |
| 4 | 0.54 | 0.67 | 0.52 | **0.75** | 0.61 |
| 5 | 0.50 | 0.47 | 0.52 | 0.53 | **0.71** |

Cross-test results among all available agents are shown in Table II. Therein, each $<i;i>$ entry (on the main diagonal) refers to the re-classification testing accuracy; note that these numbers are not even close to 1 (suggesting an inherently hard problem) and that the relative standard deviation (RSD) is 0.1049 (indicating small differences in how each agent re-classifies the training data generated by him/herself and concurring with Fig. 1, where the agent attribute seems to dominate the differences in how a problem is dealt with). Each $<i;j>$ entry describes an experiment where data for the *i*-th agent is used to build a decision tree which is subsequently tested on data for the *j*-th agent.

### B. Cost-sensitive decision trees

A cost-matrix that captures the differences between misclassifying customer complaints as requiring On Site Support vs Remote Resolution Service is:

$$C_M = \begin{bmatrix} 0 & OPESX_{OS} \\ OPESX_{RS} & 0 \end{bmatrix} \quad (1)$$

This is not a conventional misclassification cost matrix [12]; we actually note the cost incurred to carry out some maintenance activity, while we choose to ignore the cost of that activity when the model correctly predicts the class label (of the maintenance activity). So, when an On Site instance is correctly classified by our model, we assign it a cost of 0, though it incurs a cost of $OPEX_{OS}$ to resolve; instead, we reserve the cost of $OPEX_{OS}$ only for those instances which our model classifies as On Site though the class label in the data set reads Remote. Cross-testing agents with the above cost perspective is shown in Table III.

TABLE III
AVERAGE MISCLASSIFICATION COST

|   | 1 | 2 | 3 | 4 | 5 |
|---|---|---|---|---|---|
| 1 | **0.1** | 0.1 | 0.1 | 0.1 | 0.2 |
| 2 | 0.5 | **0.3** | 0.4 | 0.5 | 0.6 |
| 3 | 0.5 | 0.3 | **0.4** | 0.5 | 0.6 |
| 4 | 0.1 | 0.1 | 0.2 | **0.1** | 0.1 |
| 5 | 0.2 | 0.1 | 0.1 | 0.1 | **0.1** |

The RSD along the main diagonal now stands at 0.7071 and raises the question why agents fare so differently when asked to classify the very data they were trained on, while also highlighting substantial cost differences between agents, when some of them are used to classify data used by other agents.

It might seem that an agent may be unfairly penalized for handling a majority of serious complaints requesting on-site maintenance. We note that complaints are assigned to agents randomly (for example, we do not pick experienced agents to deal with frustrated customers); as a result, one expects that, on average, hard and easy problems are uniformly allocated. Moreover, any performance review will no doubt also focus on how specific complaints were dealt with; our technique at this point mainly serves to highlight inconsistencies in dealing with customer problems and not the actual cost of the maintenance approach (as a result, the question of why two particular agents performed a different classification on a similar set of complaints is of relatively less importance).

So, incorporating the cost aspect into the decision tree classifier allowed us to build agent-aligned models and show how these scale up to new data sets. To stress how traditional techniques provide rather just crude clues, note that traditional KPI-based agent monitoring usually consists of at least 2 indices per agent. Table IV reports on the conventional cost index for each agent (using actual values for two real KPI indices; their names as well as the formula to calculate the composite cost index are withheld due to commercial secrecy but, as most KPI indices, they do bear some relation with the effectiveness of complaint resolution, also taking into account recurring complaints). We now observe an RSD of a mere 0.0219; this clearly suggests that traditional indices grossly hide differences.

TABLE IV
TRADITIONAL KPIS AND COST EVALUATION

| Agent$_i$ | KPI$_1$ | KPI$_2$ | AGENT$_{COST}$ |
|---|---|---|---|
| 1 | 0.30 | 0.44 | 0.72 |
| 2 | 0.20 | 0.42 | 0.69 |
| 3 | 0.14 | 0.39 | 0.69 |
| 4 | 0.25 | 0.46 | 0.68 |
| 5 | 0.25 | 0.44 | 0.69 |

Table V finally reviews our experiments (ordered by increasing ability to tell agents apart in terms of performance).

TABLE V
COMPARISON OF TECHNIQUE RESOLUTION

| Table | | RSD |
|---|---|---|
| 4 | Cost using Traditional KPIs | 0.022 |
| 2 | Correctly Classified Instances only | 0.105 |
| 3 | Average Misclassification Cost | 0.707 |

### C. Advisor Flow Graph: just-in-time collaborative training

Since some agents' models seem to be better suited when classifying hereto un-seen instances, it might be reasonable to ask those agents to serve as advisors for their colleagues. Though organizational level training involves the review of past cases as well as helpdesk tactics (how to use one's communication skills, for example), the helpdesk data analysis models can serve as on-the-spot training; when an agent addresses a troubleshooting report, it makes sense to offer him/her an alternative approach to the same problem. It is crucial to present that advice as a hint and not as a recommendation to be followed; since the actual classification problem is very difficult on its own, any advice that sounds too firm might be easily resisted on grounds of just a counterexample (and, as we have seen, these do occur).

To capture a picture of which agent can help a colleague, we draw an advisor flow graph, where at source nodes we place those agents with "better" models, whereas agents who seem to act with increased misclassification costs are placed at destination nodes. Table III is used to build the advisor flow graph shown in Fig. 2, using Equation 2 for the corresponding adjacency and weight matrix. The semantics of the advisor flow graph are straightforward, with larger edge weights reflecting larger quality differences in the corresponding cross-tested models, according to Equation 2. For clarity purposes, we have drawn the graph in three levels to denote that the top level nodes are source-only nodes whereas the bottom level nodes are destination-only nodes. For our example, Agent 3 should consult Agents 1, 4 or 5, before deciding, whereas Agent 2 might also consult Agent 3, should the others be unavailable (a pop-up window describing an alternative action and the justification could suffice).

$$\begin{cases} C_{i,j} > C_{j,i} \rightarrow w(i,j) = |C_{i,j} - C_{j,i}| \\ C_{i,j} < C_{j,i} \rightarrow w(j,i) = |C_{i,j} - C_{j,i}| \end{cases} \quad (2)$$

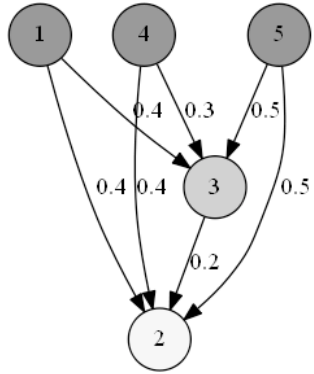

Fig. 2. An advisor flow graph showing the flow of information.

### III. CONCLUSIONS AND FUTURE WORK

We identify helpdesk agents capable of on-the-job advising their colleagues, using cost-sensitive decision tree models of how agents deal with customer complaints. These decision trees pinpoint agent differences and help build a graph to capture possible advice flows between agents. These advice flows are, essentially, by-products of standard data mining activities and can be used as non-intrusive on-the-job recommendation, suitable not only for decision making but also for reflection.

An advisor graph is a dynamic knowledge hierarchy [13] and captures the implicit reputation enjoyed by members of a community of practice, even though such knowledge may be implicit (and volatile). This is relevant to leader selection via clustering as performed in swarm-based optimization problems [14], where leaders help balance quality and diversity; in our case, this refers to organizational goals of effectiveness and efficiency. An advisor graph can gracefully scale up with the number of agents as its calculation can be carried out off-line; subsequently, selecting who-advises-whom is performed by selecting any of the available source nodes for a particular destination node. It can also scale up with the number of attributes recorded for each complaint; it is not unnatural to enhance each record with some agent-specific attributes so that agent similarities may be detected and exploited across a variety of other criteria (note, however, that recording data such as agent age, sex, education background, etc. can ran contrary to law and/or organizational regulations, which is why, for the context of our case study, we have limited data collection to the business-specific domain only).

Knowledge volatility issues arise as the advisor graph changes over time, due to the variability of the troubleshooting activities or to the helpdesk agents. We expect that, for an application, one might decide to analyze medium-sized agent groups and a larger time window. Establishing the right combination of group size and time window will likely be a difficult problem and, for all practical purposed, should probably be tackled on a trial-and-error basis.

Simply put, the contribution of this article is the use of decision trees as a classification mechanism and as a knowledge transfer mechanism for helpdesk technicians. But, while it may be easy to build the advisor graph, deploying it and monitoring how it evolves and how it actually improves quality of service is a key part of our agenda to close the loop between monitoring and acting, via peer-training.

ACKNOWLEDGMENT

The work was carried out while the first author was with the Hellenic Telecommunications Organization (OTE S.A.), Greece. We acknowledge the granting of permission to use individual technicians' data. All data is confidential and the property of OTE S.A.